\documentclass[letterpaper, 10 pt, conference]{ieeeconf}  

\IEEEoverridecommandlockouts                              

\usepackage{graphicx} 
\usepackage[linesnumbered,ruled,vlined]{algorithm2e}
\usepackage{algpseudocode}
\usepackage{amsmath} 
\usepackage{amssymb}  
\usepackage{booktabs}   
\usepackage{multirow}   
\usepackage{cuted}   
\usepackage{lipsum}
\usepackage{caption}
\usepackage{subcaption}

\title{\LARGE \bf
SHIELD: Spherical-Projection Hybrid-Frontier Integration for Efficient LiDAR-based Drone Exploration
}

\author{Liangtao Feng$^{1}$, Zhenchang Liu$^{2}$, Feng Zhang$^{3}$ and Xuefeng Ren$^{4}$
\thanks{*This work was supported by Zhuoyi Intelligent Tech Co, Ltd.}
\thanks{$^{1}$ Liangtao Feng
        {\tt\small liangtao.feng@foxmail.com}}
\thanks{$^{2}$ Zhenchang Liu
        {\tt\small liuzhenchang.2019@tsinghua.org.cn}}%
}

\begin{document}

\maketitle
\thispagestyle{empty}
\pagestyle{empty}


\begin{strip}
    \centering
    \includegraphics[width=1.0\textwidth]{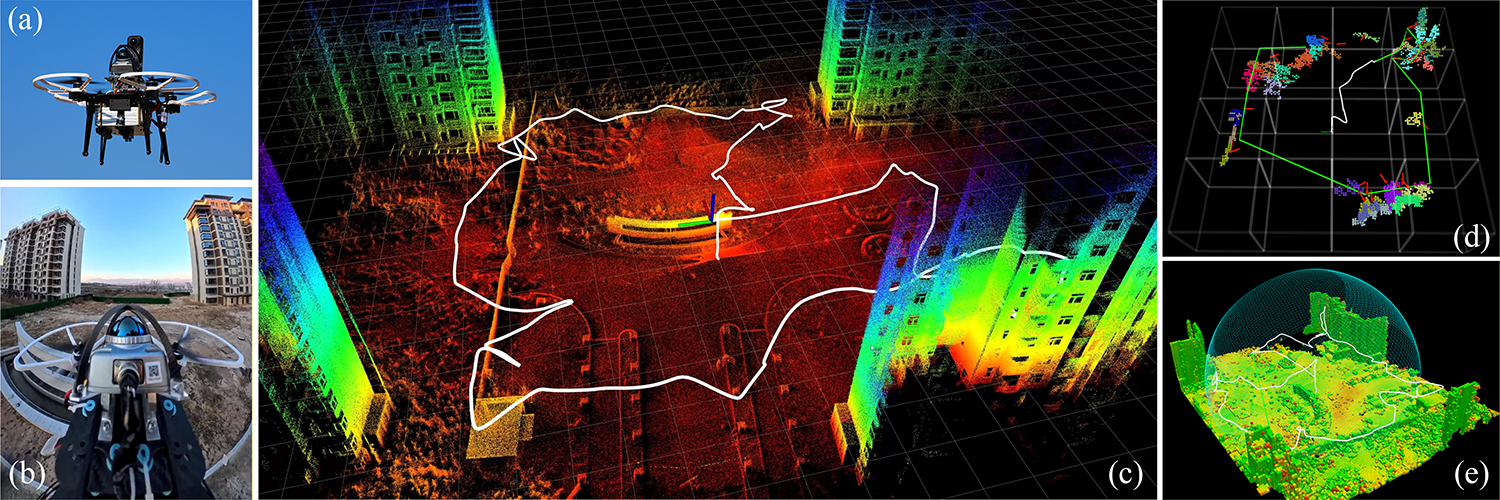}
    \captionof{figure}{Outdoor open area exploration of SHIELD (a) UAV (b) First-person view (c) Point cloud map and trajectory (d) Hgrid and frontier distribution (e) Voxel map and spherical projection}
    \label{Fig0}
\end{strip}

\begin{abstract}

This paper introduces SHIELD, a Spherical-Projection Hybrid-Frontier Integration for Efficient
LiDAR-based Drone exploration method. Although laser LiDAR offers the advantage of a wide field of view, its application in UAV exploration still faces several challenges. The observation quality of LiDAR point clouds is generally inferior to that of depth cameras. Traditional frontier methods based on known and unknown regions impose a heavy computational burden, especially when handling the wide field of view of LiDAR. In addition, regions without point cloud are also difficult to classify as free space through raycasting.  To address these problems, the SHIELD is proposed. It maintains an observation-quality occupancy map and performs ray-casting on this map to address the issue of inconsistent point-cloud quality during exploration. A hybrid frontier method is used to tackle both the computational burden and the limitations of point-cloud quality exploration. In addition, an outward spherical-projection ray-casting strategy is proposed to jointly ensure flight safety and exploration efficiency in open areas.  Simulations and flight experiments prove the effectiveness of SHIELD. This work will be open-sourced to contribute to the research community.

\end{abstract}

\section{INTRODUCTION}

In recent years, exploration methods based on unmanned aerial vehicles (UAVs) have attracted significant research interest. A central challenge in UAV exploration is reliable perception and environment representation, which largely depends on the types of onboard sensors and the corresponding algorithms for mapping and planning. 

One common approach is vision-based exploration, which typically employs the depth camera to perceive the environment\cite{c1}\cite{c2}. Depth cameras can directly capture 3D structure and enable dense mapping using visual odometry, making them suitable for structured indoor and outdoor scenes with sufficient texture and lighting conditions. However, the field-of-view (FOV) of the depth camera is relatively small. Depth-camera-based visual localization provides accurate short-range geometric perception and metric-scale estimation, but its limited sensing range and sensitivity to illumination and motion significantly constrain its robustness in high-speed and large-scale UAV exploration. RACER\cite{c3} employs UWB for localization, while EDEN\cite{c4} relies on LiDAR-based odometry for state estimation.

Another common method is LiDAR-based exploration, which has garnered increasing attention. Although LiDAR provides a wide FOV and information-rich point clouds, it also introduces several challenges. 

First, the wide FOV inherently leads to non‐uniform observation quality. The sparsity of the point cloud varies as a function of distance and angle. Traditional frontier-based exploration methods typically rely on the global uniform occupancy map\cite{c9}, and the various extensions\cite{c12}\cite{c14}. Star-searcher\cite{c13} builds an occupancy map with the observation distances of voxels. EPIC\cite{c10} builds an observation map that represents surface observation quality. 

Second, the wide FOV makes the selection of frontiers different. For example, consider the Livox Mid‑360 LiDAR, which features a full 360° horizontal field of view and a vertical field spanning from -7° to +52°. Common free-unknown frontiers can cover a very large area and appear in large numbers, increasing the burden of exploration. FLARE \cite{c5} utilizes unknown regions for guidance. SOAR\cite{c6} adopts a surface frontier-based exploration strategy to acquire the surface geometry rapidly. A multi-resolution frontier-based planner is designed for 3D exploration, to low the computational burden\cite{c7}. Some exploration methods abandon frontier-based strategies in favor of sampling-based approaches, such as TARE\cite{c8} and \cite{c15}.

Third, LiDAR does not generate depth information at a uniform and fixed resolution across its FOV, in contrast to depth cameras. It produces returns only in the directions where obstacles are present. When no obstacles exist, the absence of returns hinders reliable assessment of free space in that direction.\cite{c11}. This means that common raycast schemes, unless operating in fully enclosed indoor environments, find it difficult to accurately distinguish between free and unknown space. EPIC\cite{c10} utilizes collision-free spheres to determine free space coverage, which cleverly addresses this problem. This implicitly assumes that regions within the LiDAR FOV with no returns are safe to fly through. However, in cases where the LiDAR is occluded by the vehicle body, this approach may lead to incorrect judgments. 

From the above investigation, it is evident that LiDAR‑based exploration differs fundamentally from exploration with depth cameras and requires greater consideration of sensor‑specific characteristics.

Though the frontier-base exploration method is widely adopted, it may suffer from inefficiencies such as frequent revisits and suboptimal global coverage when used in isolation\cite{c16}. Global guidance methods are introduced to consider the visiting order of frontiers\cite{c17}\cite{c18}. However, such a method focuses on the known regions and overlooks the unknown regions, thus reducing the overall exploration efficiency. The coverage path (CP) algorithms aim to cover the area with minimum overlapping\cite{c19}. In RACER\cite{c3}, exploration tasks for multiple UAVs are assigned using a coverage path planning method. To reduce the high online computational burden, the topological graph is incorporated into the CP method \cite{c2}. SphereMap\cite{c20} builds a topological map based on collision-free sphere connectivity.

To better address these problems, this work proposes SHIELD, Spherical-Projection Hybrid-Frontier Integration for Efficient LiDAR-based Drone Exploration, as shown in Fig.\ref{Fig0}. 

The main contributions of this work are:
\begin{enumerate}
\item An observation-quality occupancy map based on surface normals is presented, along with a corresponding quality-based raycast method.
\item A hybrid frontier-based exploration strategy is developed for LiDAR-based exploration.
\item An outward spherical-projection raycast strategy, together with a calibration method, is proposed to assess free space in the absence of sensor returns.
\end{enumerate}

This work will be open-sourced to contribute to the research community.

\section{Problem Definition and System Overview}
The problem addressed in this work is exploring a large-scale, unknown, open, bounded 3D area $A \in \mathbb{R}^3$ using Autonomous aerial vehicle (AAV). Additionally, the proposed method is able to explore in confined spaces, narrow passages, and environments with irregular geometry.  After exploration, the system generates a point cloud map that reconstructs the 3D structure, as well as an occupancy map that represents the free, occupied, and unknown regions of space.

\begin{figure}[h]
    \centering
    \includegraphics[width=0.4\textwidth]{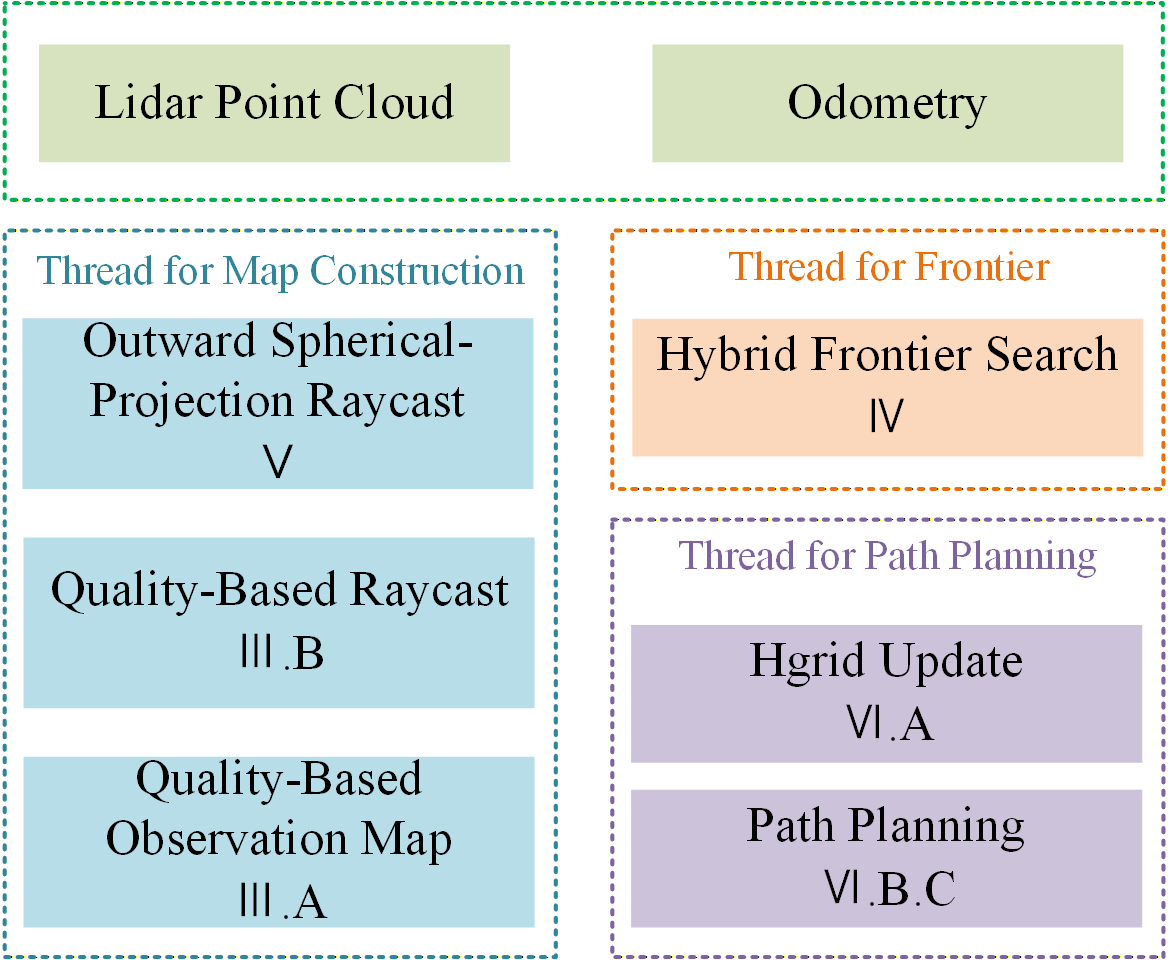}
    \caption{Structure of SHIELD}
    \label{Fig20}
\end{figure}

The system structure is represented in Fig.\ref{Fig20}. This work employs a multithreading mechanism to balance the load among various tasks and improve runtime efficiency.

\section{Quality-Based Oberservation Map and Raycast}
The traditional occupancy grid map relies on its probabilistic representation of occupancy. It updates occupancy information by casting rays from detected points to the sensor origin, marking traversed voxels as free and start points as occupied. The probabilistic model allows for the integration of noisy sensor measurements over time, refining the occupancy estimates as more data becomes available. When the voxel size becomes big and the FOV is large, the raycast may incorrectly set the occupied voxel as free. Such errors will directly affect the flight-path planning and in turn reduce exploration efficiency.
\begin{figure}[h]
    \centering
    \includegraphics[width=0.4\textwidth]{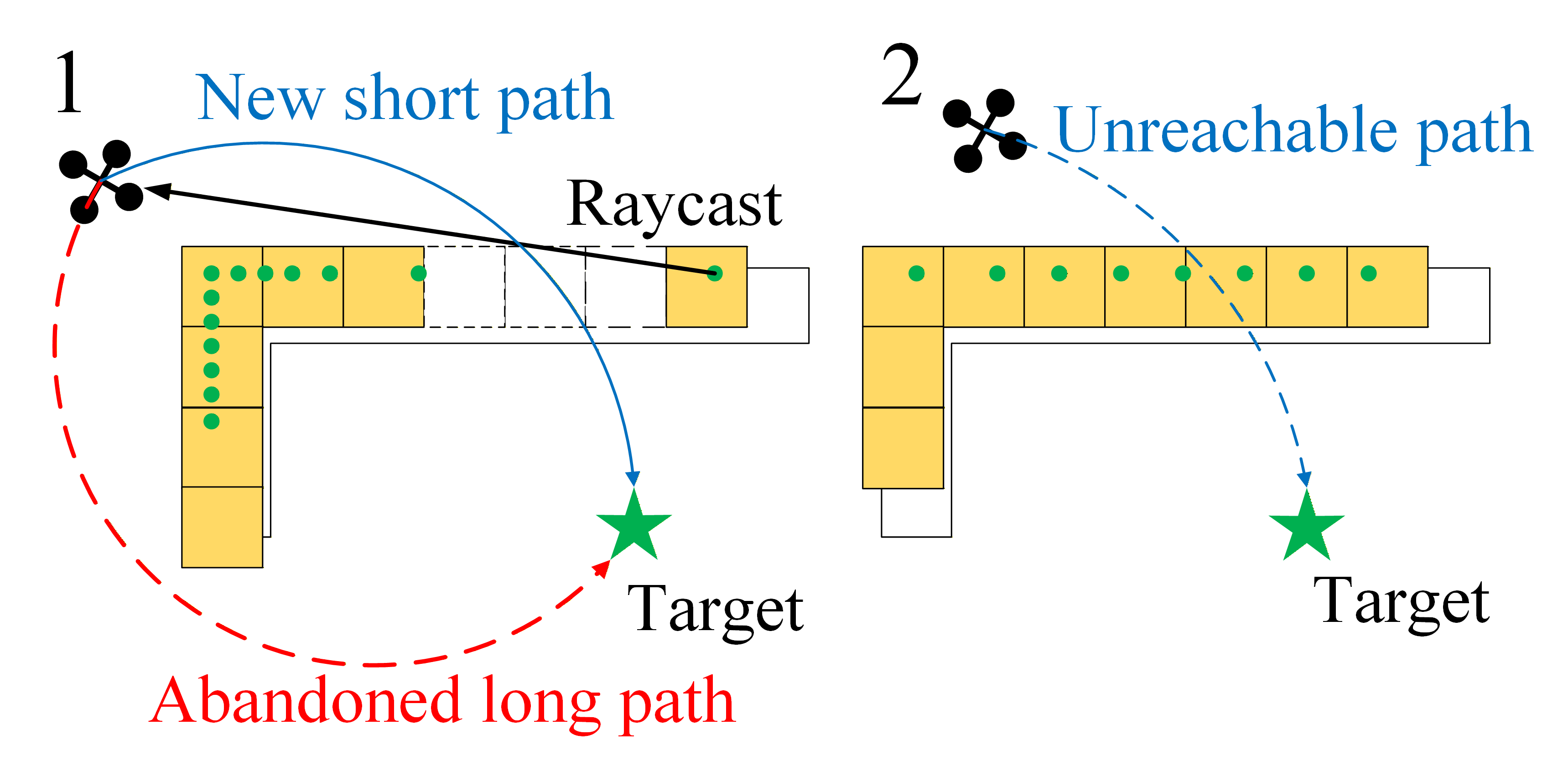}
    \caption{Wrong raycast under big FOV}
    \label{Fig1}
\end{figure}
As shown in Fig. \ref{Fig1}, when the UAV is at Position 1, the rays incorrectly pass through nearby voxels and cause them to be falsely marked as free, leading to a new shorter path. However, when the vehicle moves to Position 2 and the observation quality improves, the previously planned path is no longer valid, thus causing loitering.

To address this, this section introduces a quality-based oberservation map $\mathcal{M}_{\text{qua}}$ and the correspoding raycast method.

\subsection{Quality-Based Oberservation Map Construction}
In EPIC\cite{c10}, observation quality is defined using the ray length and angular separation of LiDAR. However, such a method implicitly assumes that obstacles are spatially continuous. Inspired by this, this work adopts a method that extracts information from the environment rather than relying solely on the LiDAR sensor(Alg.\ref{alg_map}). 

\begin{figure}[h]
    \centering
    \includegraphics[width=0.3\textwidth]{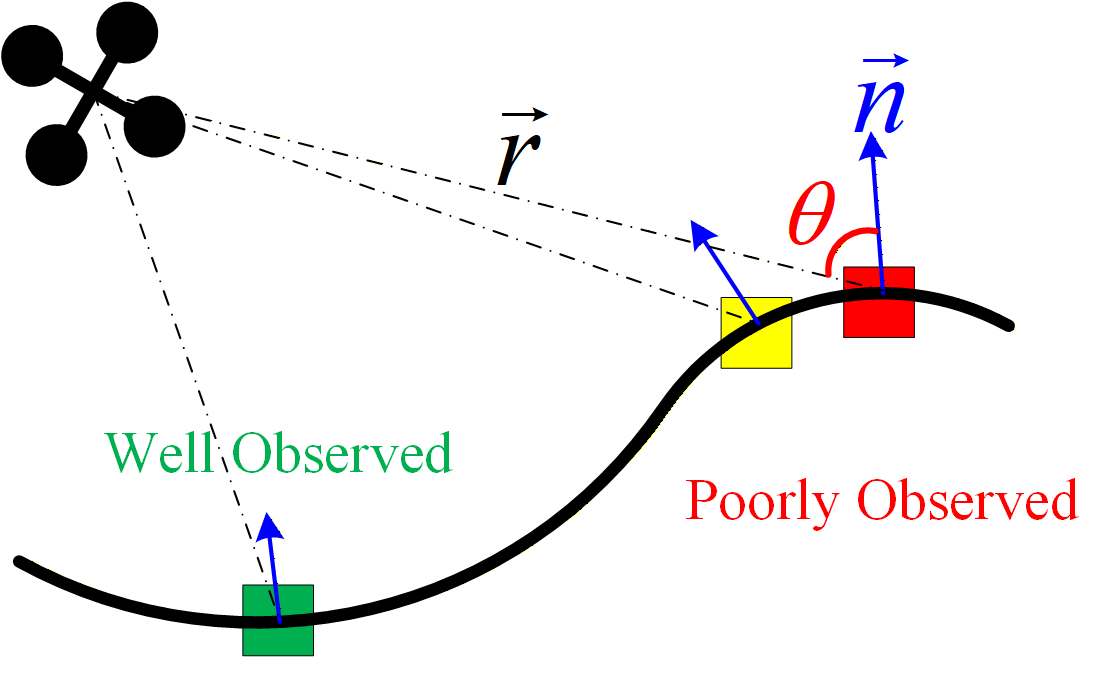}
    \caption{Observation of point cloud quality}
    \label{Fig2}
\end{figure}

In each cycle, the current point cloud under the global frame ($\mathcal{P}_{global}$) is input. First, the point clouds from the last several cycles are merged to produce a denser point cloud ($\mathcal{P}_{buffer}$). Second, the point number $\mathcal{N}$ in the corresponding voxel is counted and stored. $\mathcal{N}$ is reset to 0 and recounted in each cycle. The surface normals of $\mathcal{P}_{buffer}$ are estimated using Point CLoud Libraby (PCL). The surface normals at the center of each voxel are stored as the feature. Meanwhile, based on the position of the UAV $\vec{x}$ and the current point position $\vec{p}$, a ray is cast and the observation angle $\theta$ between the ray $\vec r$ and the surface normal $\vec n$ is computed as Equation (\ref{eq1}). This process is illustrated in Fig.\ref{Fig2}. The biggest cosine value of $\theta$ (smallest angle, normal to the obstacle) in each voxel is stored as the observation quality $\mathcal{Q}$.

\begin{equation}
\mathcal{Q} = \cos{\theta} =\left|\frac{\vec{r}\cdot\vec{n}}{\left\|\vec{r}\right\| \left\|\vec{n}\right\|}\right|
\label{eq1}
\end{equation}

\begin{algorithm}
\caption{Oberservation Map Construction}\label{alg_map}

\KwIn{$\mathcal{P}_{global}$}
\KwOut{$\mathcal{N}$, $\vec n$, $\mathcal{Q}$}

$\mathcal{P}_{buffer}$ $\leftarrow$ \textbf{MergePointCloud}($\mathcal{P}_{global}$)\;
$\mathcal{N}$ $\leftarrow$ \textbf{CountPointNum}($\mathcal{P}_{buffer}$)\;
$\vec n$ $\leftarrow$ \textbf{NoramEstimation}($\mathcal{P}_{buffer}$)\;
\For{$p \in \mathcal{P}_{global}$}{
    $\vec r$ $\leftarrow$ \textbf{CalRay}($\vec{x}$, $\vec{p}$) \;
    $\mathcal{Q}$ $\leftarrow$ \textbf{CalQuality}($\vec r$, $\vec n$)\;
    
}
\textbf{StoreInVoxel}($\mathcal{N}$, $\vec n$, $\mathcal{Q}$)\;
\end{algorithm}

\subsection{Raycast based on Quality}

The raycast is considered based on the following factors: ray quality, the quality of historical observations, and the number of past point clouds, which are obtained and stored in the observation map.  First, At the ray’s origin, compute the observation angle $\theta$. If $\theta$ is sufficiently small, classify the ray as a 'well-observed' ray. The corresponding voxel is marked as 'hit'. Second, as the ray traverses the voxel grid, if a voxel is marked as occupied, retrieve the associated surface normal vector $\vec n$  and the number of historical points $\mathcal{N}$ for that voxel. Third, the voxel is marked as 'miss' if the following conditions are met:
\begin{enumerate}
\item The ray is a “well-observed” ray.
\item The number of historical points is less than a threshold $\mathcal{N}_{\text{th}}$.
\item The observation angle $\theta$ between ray and the normal vector of the current voxel is smaller than a threshold $\theta_{\text{th}}$.
\item The observation quality $\mathcal{Q}$ of the current voxel is less than a threshold $\mathcal{Q}_{\text{th}}$
\end{enumerate}

\section{Hybrid-Frontier Strategy}

\subsection{Frontier Detection and Clustering}

In most exploration algorithms, a “frontier” is defined as the boundary between free and unknown space. However, for LiDAR sensors with a large FOV, the regions of such a boundary become so large that this definition imposes heavy computational burden. EPIC\cite{c10} selects voxels that lie between well-observed and poorly-observed regions as frontiers. However, most frontiers of EPIC lie on the ground when it comes to the large-scale outdoor scene, thus reducing exploration efficiency.

\begin{figure}[h]
    \centering
    \includegraphics[width=0.45\textwidth]{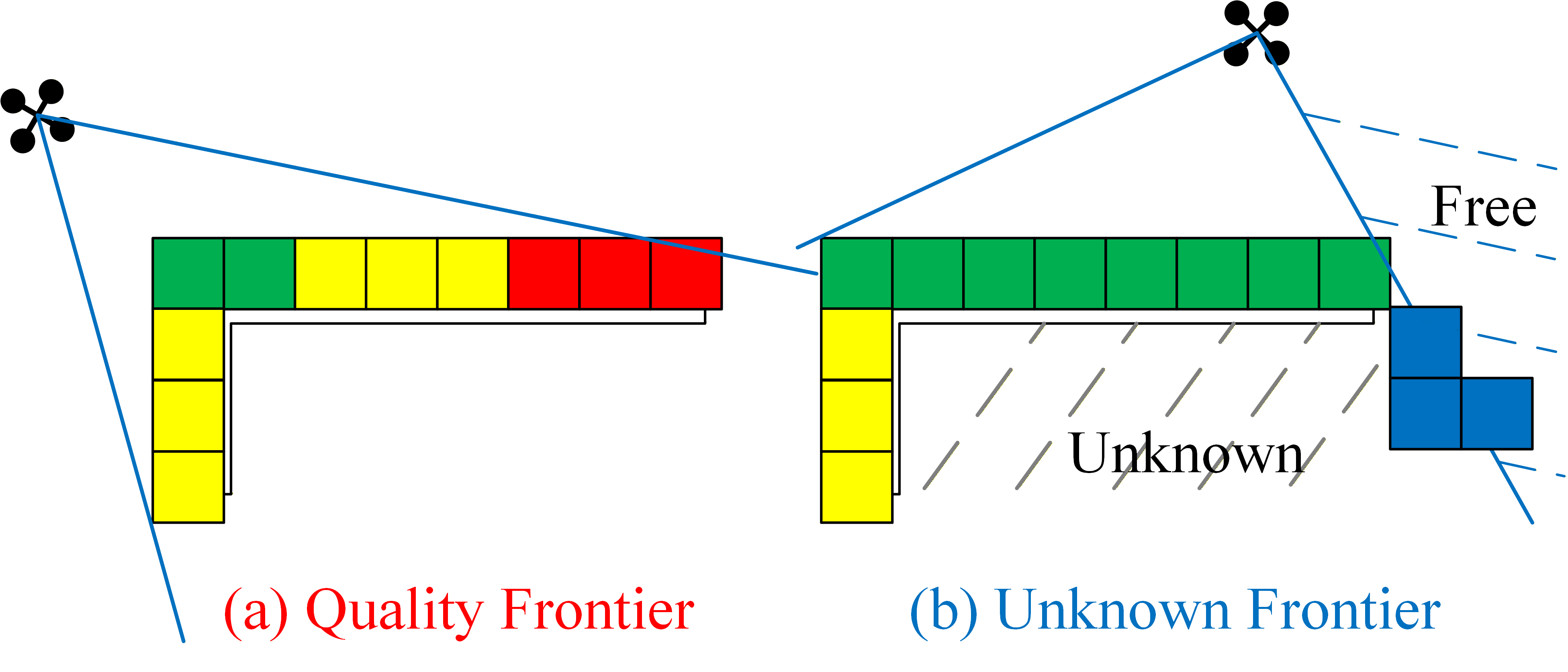}
    \caption{Definition of the Hybrid Frontier}
    \label{Fig3}
\end{figure}

To balance exploration efficiency, observation quality, and computational load, this paper proposes a hybrid frontier strategy. The frontier consists of two types: quality frontier (Fig.\ref{Fig3}a) and unknown frontier(Fig.\ref{Fig3}b).

During the update process of $\mathcal{M}_{\text{qua}}$, the voxels with poor observation quality ($\mathcal{Q}<\mathcal{Q}_{\text{th}}$) are selected as a set $\mathcal{S}_{\text{qua}}$. In $\mathcal{S}_{\text{qua}}$, voxels are grouped into clusters according to a Euclidean-distance threshold and the normal vector. The final selected clusters have a certain minimum size and are nearly planar. These clusters define the candidate quality frontier $\mathcal{F}_{\text{qua}}$, waiting for checking whether appropriate viewpoints can be generated.

To serve a role similar to that of the “free–unknown frontier”, voxels, which lie in free space and are adjacent to both unknown and well‑observed occupied voxels, are selected as $\mathcal{S}_{\text{unknown}}$. If a cluster contains at least a minimum number of voxels, it is defined as the candidate unknown frontier $\mathcal{F}_{\text{unkown}}$.

\subsection{Viewpoint Generation}

For a frontier cluster $\mathcal{F}$, candidate viewpoints are uniformly sampled in a cylindrical coordinate system centered at its center. For each such viewpoint, the yaw angle is oriented toward the center of the cluster.
In the case of the quality frontier $\mathcal{F}_{\text{qua}}$, the surface normal information previously estimated for the voxels in $\mathcal{F}_{\text{qua}}$ is employed to compute an observation-quality metric. For each voxel $v\in \mathcal{F}_{\text{qua}}$, the angle between the ray from viewpoint $z$ and the surface normal $\vec{n}$ is evaluated. The average observation quality $Q(z)$ under viewpoint $z$ is then computed as Equation (\ref{eq232}). The $z$ with the highest average quality is selected as the final viewpoint of $\mathcal{F}_{\text{qua}}$.

\begin{equation}
Q(z) = \frac{1}{\left|\mathcal{F}_{\text{qua}}\right|}\sum\limits_{\mathcal{F}_{\text{qua}}}{Q(z)}
\label{eq232}
\end{equation}

In the case of the candidate unknown frontier $\mathcal{F}_{\text{unknown}}$, the current occupancy map is consulted to determine the voxels $\mathcal{F}_{\text{vis}} $ whose occupancy state remains unknown and are in FOV of the viewpoint. The candidate viewpoint with the largest number of $\mathcal{F}_{\text{vis}}$ is selected as the cluster’s final viewpoint.

After clustering voxels and generating the viewpoints, data is stored for global path planning. If more than 20 percent of the voxels in a frontier fail to meet the required condition, the frontier is discarded, and its voxels are re-clustered.

\section{Outward Spherical-Projection Raycast}

\begin{figure}[h]
    \centering
    \includegraphics[width=0.4\textwidth]{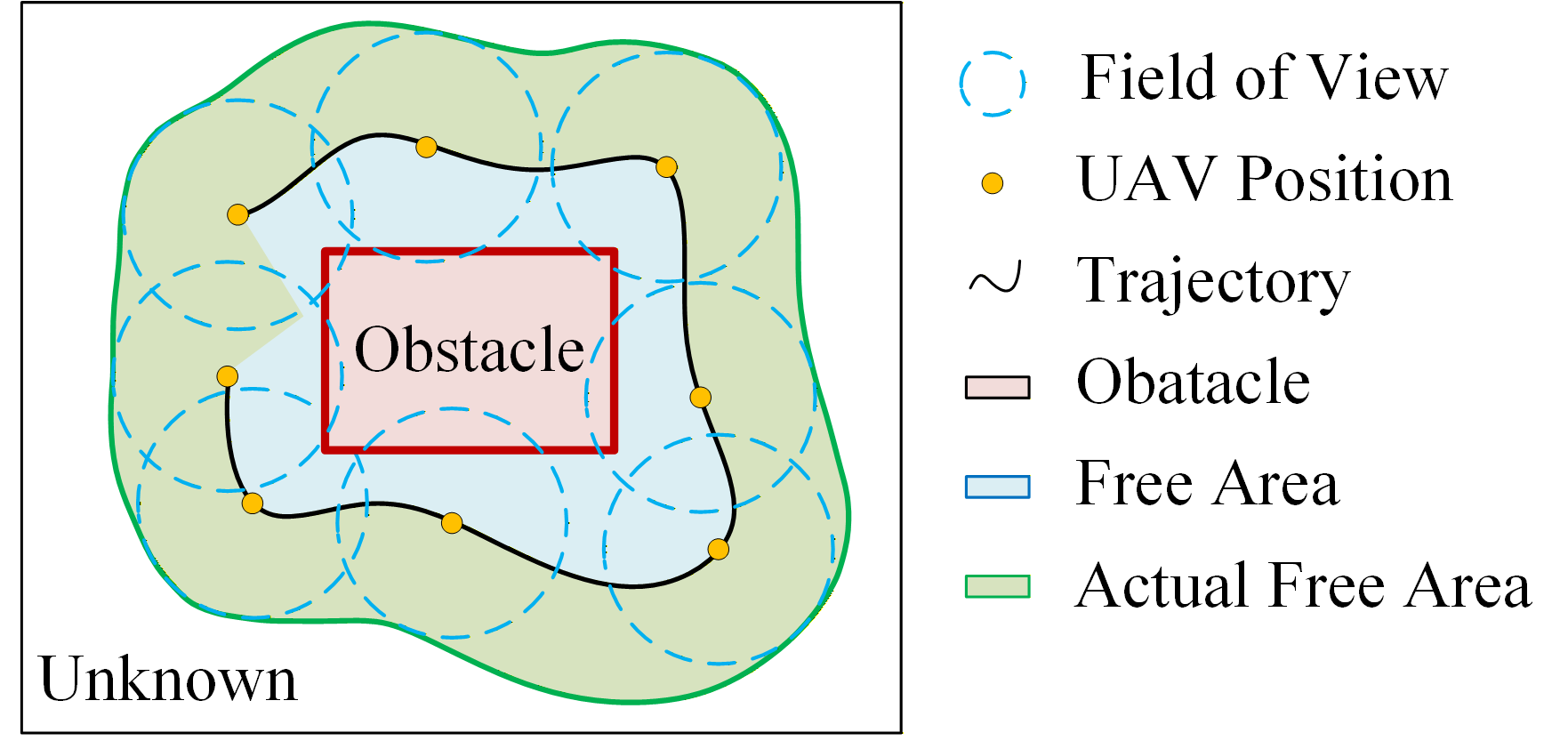}
    \caption{Definition of the Hybrid Frontier}
    \label{Fig11}
\end{figure}

As shown in Fig.\ref{Fig11}, when the UAV comes to a large-scale outdoor scene with few obstacles, the regions capable of reflecting LiDAR returns, thus returning point clouds, are drastically reduced. The regions without point clouds are indeed safe to fly (green area), but cannot be marked as free in $\mathcal{M}_{\text{qua}}$ (blue area). This also complicates trajectory planning and limits the set of feasible viewpoints, which in turn reduces the efficiency of exploration.

In this paper, an outward spherical-projection raycast method is proposed, to address the problem of marking free voxels in the regions without point clouds.

\subsection{Outward Spherical-Projection Raycast}

\begin{figure}[h]
    \centering
    \includegraphics[width=0.5\textwidth]{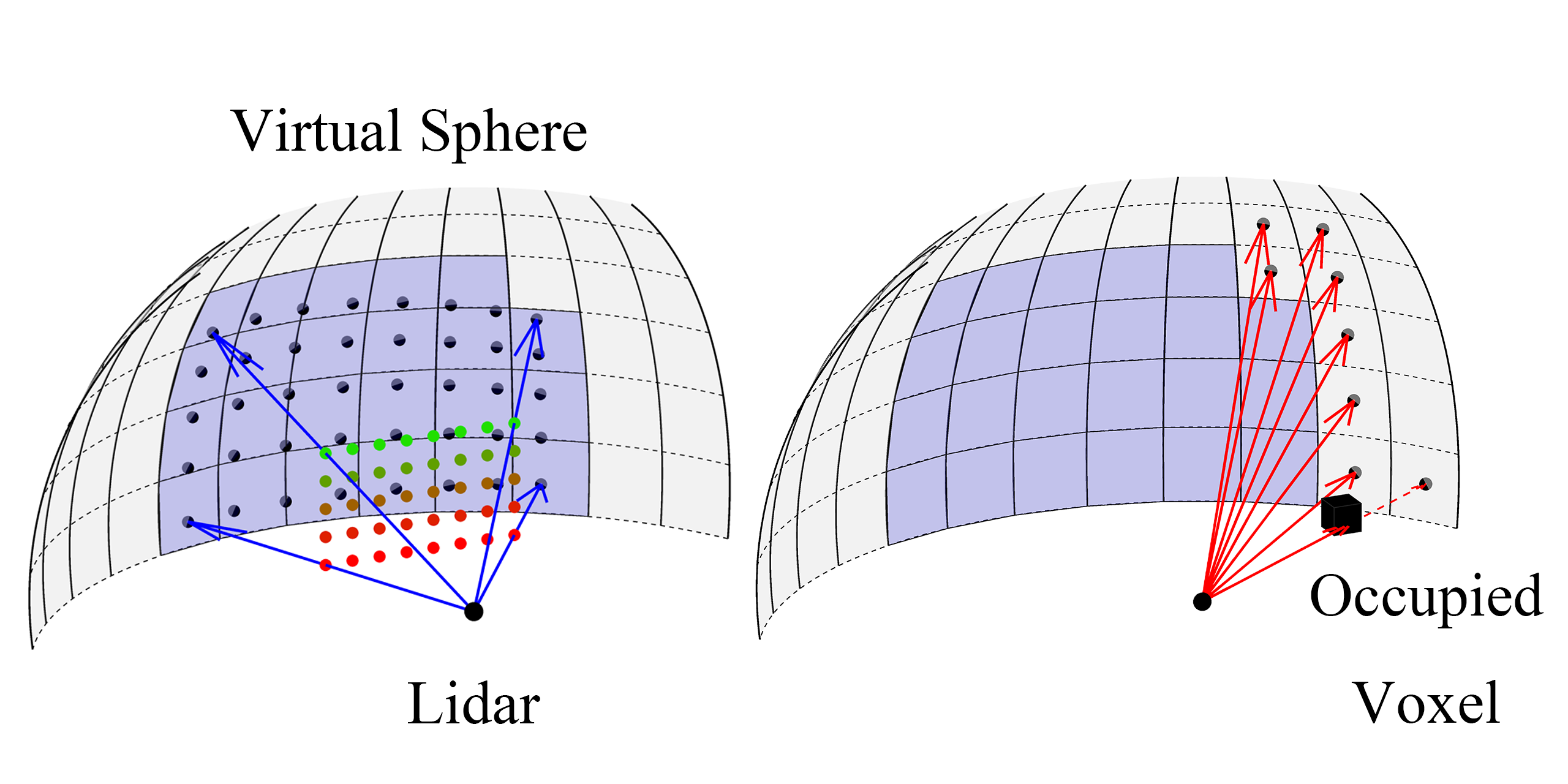}
    \caption{Outward Spherical-Projection Raycast}
    \label{Fig4}
\end{figure}

The process is shown in Fig. \ref{Fig4}. First, the FOV range of LiDAR sensor is abstracted as a virtual spherical surface. Taking the Mid360 as an example, its FOV can be represented as a spherical surface with a radius of 30 meters, spanning 360 degrees in longitude and from -7 to 52 degrees in latitude. The spherical surface is divided into grid cells of equal longitude and latitude intervals. Then, the point clouds $(x_{\text{i}},y_{\text{i}},z_{\text{i}})$ under the LiDAR frame is projected onto the virtual spherical surface (blue arrow) according to its longitude $\phi_{\text{i}}$ and latitude $\phi_{\text{i}}$, as Equation (\ref{eq2}), and the number of point clouds falling into each grid cell is counted. A grid cell is considered unobstructed in its corresponding direction if it contains no points and its surrounding neighborhood within a certain range is also empty. Finally, convert the centers of the unobstructed cells from the LiDAR frame to the global frame, and raycast from the UAV position to these centers (red arrow). Voxels traversed by the ray are marked as "miss". However, once the ray encounters an occupied voxel, it is terminated, and no voxels along its path are marked.
\begin{equation}
\begin{split}
r_{\text{i}} = \sqrt{x_{\text{i}}^2+y_{\text{i}}^2+z_{\text{i}}^2}\\
\phi_{\text{i}} = \arctan{\frac{y_{\text{i}}}{x_{\text{i}}}}\\
\theta_{\text{i}} = \arcsin{\frac{z_{\text{i}}}{r_{\text{i}}}}
\end{split}
\label{eq2}
\end{equation}

\subsection{Self‑Calibration of Spherical Projection}

Radar sensors are frequently occluded by nearby structures (e.g., protective guards or the vehicle body), which typically fall within the LiDAR's minimum detectable range. Without proper handling, the lack of point clouds from these directions results in incorrect "free" voxels in the map.

To address this, this work proposes a self-calibration method for marking areas devoid of point clouds. The procedure begins in an enclosed space with rich reflectivity, where the UAV and LiDAR are agitated to cover a wide range of orientations. This maximizes the likelihood of obtaining LiDAR point clouds from most directions. Consequently, any grid that persistently lacks point clouds can be reliably attributed to close-range occlusion, rather than to free space.

Throughout this procedure, the latitude and longitude of every point cloud are computed, and the density of points per grid cell on the sphere is evaluated. Meanwhile, by integrating the intensity information $I_{\text{i}}$ of the points, the weighted average position $(\overline{\theta},\overline{\phi})$ is derived in each grid, as Equation (\ref{eq5}).
\begin{equation}
\begin{split}
\overline{\theta} = \sum\limits_{\text{i}}^{\text{grid}}{\theta_{\text{i}}I_{\text{i}}}, 
\overline{\phi} = \sum\limits_{\text{i}}^{\text{grid}}{\phi_{\text{i}}I_{\text{i}}}
\end{split}
\label{eq5}
\end{equation}

The grid is classified as occluded (and thus excluded from subsequent ray casting) when it meets both criteria:
\begin{enumerate}
\item Its point number falls below a threshold.
\item The weighted average location from the grid center is more than a set tolerance $\lambda_{\text{th}}$, as Equation (\ref{eq7}).
\end{enumerate}
\begin{equation}
\sqrt{(\overline{\phi}-\phi_{\text{center}})^2+(\overline{\theta}-\theta_{\text{center}})^2} > \lambda_{\text{th}}
\label{eq7}
\end{equation}

\begin{figure}[h]
    \centering
    \begin{subfigure}{0.235\textwidth}
        \centering
        \includegraphics[width=\textwidth]{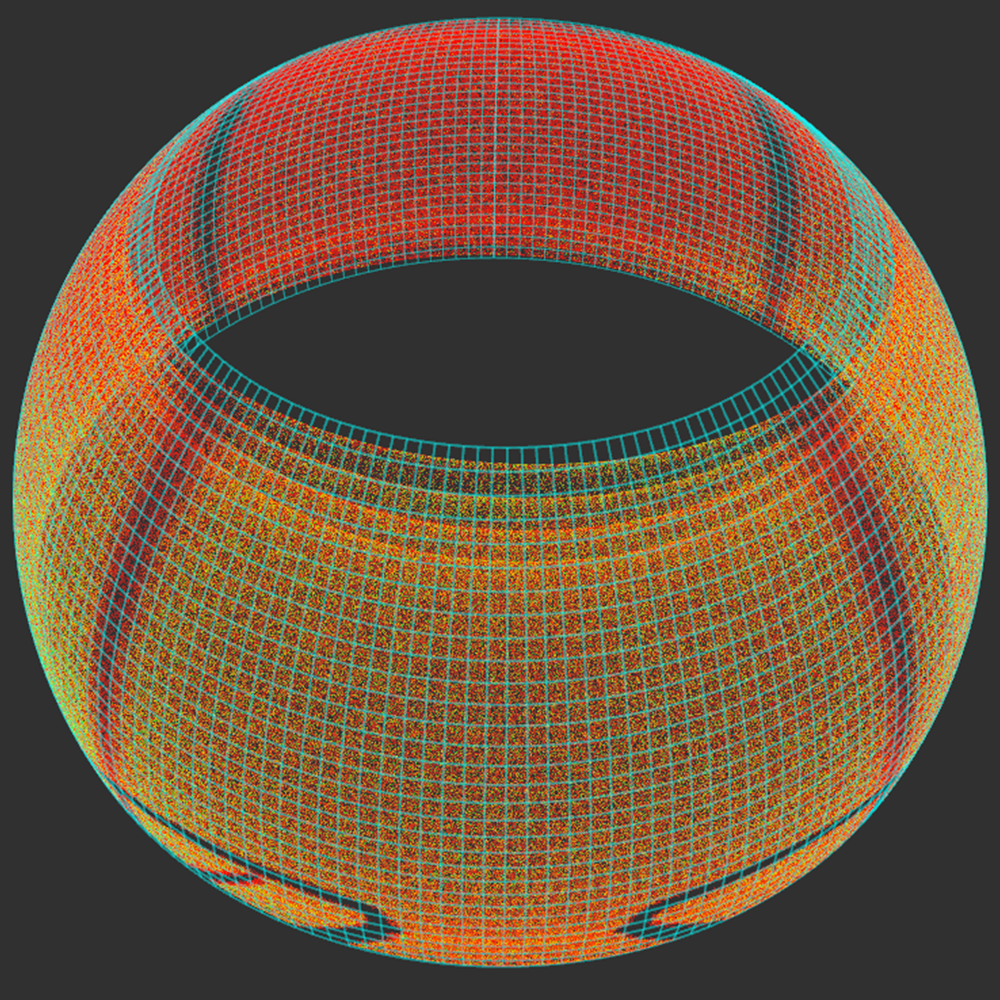}
        \caption{Point clouds projected to the virtual sphere}
        \label{fig:sub1}
    \end{subfigure}
    \hfill
    \begin{subfigure}{0.235\textwidth}
        \centering
        \includegraphics[width=\textwidth]{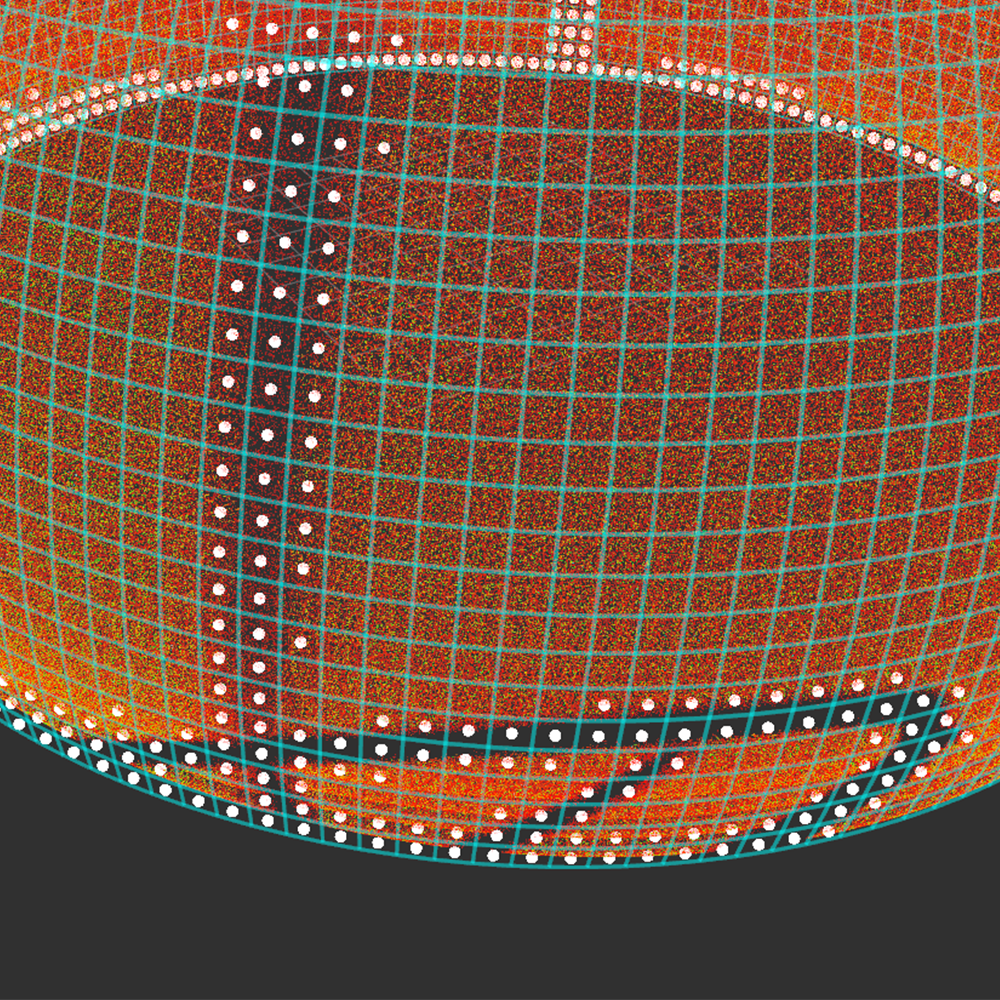}
        \caption{Occluded grids of the virtual sphere}
        \label{fig:sub2}
    \end{subfigure}
    \caption{Self-Calibration of Spherical Projection}
    \label{Fig5}
\end{figure}

Fig.\ref{Fig5} shows the result of calibration. In Fig.\ref{Fig5}a, the shadows on the spherical surface are caused by occlusions from the rotor guards and LiDAR protective rings. These regions are correctly identified in Fig.\ref{Fig5}b and are not subjected to outward raycasting.

The calibration process takes approximately 20–30 seconds. Moreover, this calibration only needs to be performed once for each assembled aircraft and can substantially improve its perception capability at minimal cost.

\section{Path Planning}

This work employs an improved coverage-path planning method to coordinate global path planning. The global path planning is formulated as an Asymmetric Traveling Salesman Problem (ATSP). The space to explore is divided into grids of same size (Hgrid). 
\subsection{Hgrid State}
The Hgrid has three states:
\begin{enumerate}
\item DORMANT: no frontiers inside, $\mathcal{N}_{\text{frt}}=0$.
\item INACTIVE: the exploration rate $\mathcal{\rho}$ reaches the threshold $\rho_{\text{th}}$(e.g., 95\%)
\item ACTIVE.
\end{enumerate}

The transition rules between the three states are described in Alg.\ref{alg_grid_update}. The dormant state for an Hgrid is introduced to account for cases in which the exploration space is irregular and either (1) the entire Hgrid is unreachable, or (2) a large portion of the Hgrid consists of unreachable regions. Moreover, the dormant grids can be reactivated when unknown frontiers occur, to avoid incomplete exploration. A grid becomes inactive once exploration in that region reaches a predefined threshold and will not be reactivated, preventing redundant revisits and improving exploration efficiency. 

\begin{algorithm}[h]
\caption{Grid State Update}\label{alg_grid_update}

\Switch{state}{
  \uCase{ACTIVE}{
    \If{$\rho > \rho_{\text{th}}$ }{
      deactivateHGrid()\
    }
    \If{$\mathcal{N}_{\text{frt}}==0$}{
      dormantHGrid()\
    }
  }
  \uCase{DORMANT}{
    \If{$\mathcal{N}_{\text{unknownfrt}} > 0$}{
      activateHGrid()\
    }
  }
  \Case{INACTIVE}{
    nothing\
  }
}

\end{algorithm}

\subsection{Hgrid Path Planning}
Path planning on the Hgrid is performed only within active grids. The visit order of Hgrid is determined by solving an ATSP, whose cost matrix is denoted by $\boldsymbol{C}_{Hgrid}$ and whose dimension is 1 + the number of active Hgrid. The cost function must be defined to get $\boldsymbol{C}_{Hgrid}$.

First, the cost from the UAV to the i-th Hgrid $C_{d,i}$ is computed. The path length $l(d,g_i)$ from the current UAV position $\vec x$ to the free center of the i-th Hgrid $\vec g_{free,i}$ is computed using an A* search algorithm. The selection of the free center as the representative of the Hgrid is predicated on the principle that the viewpoints must be generated in free space. Besides, the cost of velocity $l_v(d,i)$ is considered to prevent hesitation, as Equation \ref{eq8}. If the distance from the UAV to this Hgrid exceeds the  three-dimensional size $\vec s$ of the Hgrid measured by its $L_{2}$ norm, which may lead to excessive search time, the path length is instead approximated by 1.5 times the Euclidean distance between the UAV and the free center $\bar{g}_{free,2}$ of the next Hgrid $\bar{H}_2$. 
\begin{equation}
\begin{split}
l_v(d,g_i) = (\vec g_{free,i} - \vec x)\cdot \frac{\vec v}{\left\|\vec v \right\|}\\
C(d,g_i) = l(d,g_i) - l_v(d,g_i)
\end{split}
\label{eq8}
\end{equation}

Second, the cost from Hgrid to Hgrid is computed as Equation \ref{eq10}. The cost from the i-th Hgrid to the j-th Hgrid is composed of two components: the distance between their free centers, and a penalty term proportional to the number $m$ of already known frontiers in the j-th Hgrid. The second component serves as a penalty term that increases with the number of known frontiers in the target Hgrid, effectively biasing the cost function toward frontier-rich regions.

\begin{equation}
C(i,j) = \left\|\vec g_{free,i} - \vec g_{free,j}\right\| - m \times \left\|\vec s \right\|
\label{eq10}
\end{equation}

Worth noting, both the velocity cost and the penalty term of frontier number are bounded by predefined limits to prevent either term from excessively dominating the distance cost in the overall cost function. The final cost matrix $\boldsymbol{C}_{cp}$ is defined as Equation (\ref{eq11}).

\begin{equation}
\boldsymbol{C}_{Hgrid}(i,j) =
\begin{cases}
0&i=j\\
C(d,g_j)& i=0\\
C(g_i,g_j)& i \neq j, i \neq 0\\
\end{cases}
\label{eq11}
\end{equation}

After solving ATSP with $\boldsymbol{C}_{cp}$, the order $\bar{\Lambda}=\{\bar{H}_1,\ldots,\bar{H}_k,\dots,\bar{H}_n\}$ of visiting Hgrid is determined.

\subsection{Coverage Path Planning}

After the order of \(H_{\text{grid}}\) is determined, the viewpoints to be explored are sorted accordingly.
Let the set of viewpoints associated with the frontiers inside the first grid $\bar{H}_1$ of $\bar{\Lambda}$ be denoted by $\zeta = \{\, z_1, \ldots, z_k, \ldots, z_m \,\}$, where each $z_k$ represents a candidate viewpoint within $\bar{H}_1$.
This selection induces a new asymmetric traveling salesman problem (ATSP) that involves three classes of nodes:
the current UAV position \(\vec{x}\), the set of frontier viewpoints $\zeta$, and the free center $\bar{g}_{\mathrm{free},2}$ of the subsequent grid $\bar{H}_2$.

To formalize the cost structure, define the cost matrix $\boldsymbol{C}_{cp} \in \mathbb{R}^{N\times N}$, where $N = 3 + |\zeta|$, as Equation (\ref{eq12}). The dimension consists of the UAV, a depot, the next Hgrid $\vec{g}_{free}$ and the viewpoints $\zeta$.
\begin{equation}
\label{eq12}
\boldsymbol{C}_{cp} = 
\begin{bmatrix}
M_{inf} & -M_{inf} & \vec{M}_{inf} & M_{inf} \\
M_{inf} & M_{inf} & \boldsymbol{C}(d,z) & M_{inf} \\
\vec{0} & \vec{0} & \boldsymbol{C}(z,z) & \boldsymbol{C}^{T}(z,g) \\
-M_{inf} & M_{inf} & \boldsymbol{C}(z,g) & M_{inf}
\end{bmatrix}
\end{equation}

$M_{inf}$ is a big positive number. $\vec{M}_{inf}$ is a vector of $\mathbb{R}^{1\times |\zeta|}$. $\boldsymbol{C}(d,z)$ represents the cost from the UAV to viewpoints in $\zeta$, which is a vector of $\mathbb{R}^{1\times |\zeta|}$. $\boldsymbol{C}(z,z)$ represents the cost between viewpoints, which is a matrix of $\mathbb{R}^{|\zeta| \times |\zeta|}$. $\boldsymbol{C}(z,g)$ represents the cost from $\bar{g}_{\mathrm{free},2}$ to viewpoints, which is $\mathbb{R}^{1\times |\zeta|}$. All the cost is computed same as $l(d,g_i)$ in Equation (\ref{eq8}), which is computed using an A* search algorithm.

After solving the ATSP with $\boldsymbol{C}_{cp}$, the visiting order of viewpoints $\bar{\zeta}$ is determined. The UAV initiates local path planning based on its current position, velocity, and the targeted viewpoint.

\subsection{Local Path Planning}

Based on the current viewpoint generated by the global planner, a B-spline position trajectory\cite{c25} that satisfies the UAV's mobility constraints is planned within the free voxels of the occupancy map. Then, the yaw trajectory is designed according to the constraints of the trajectory duration and the viewpoint yaw angle. Since the UAV flies within free voxels, flight safety is guaranteed, and there is no need to constrain the field of view according to the movement direction. Therefore, based on the UAV’s maximum yaw rate, priority is given to determining whether the rotation of the UAV to the viewpoint yaw angle can satisfy the time constraints. If the constraints are met, the yaw angle of the UAV in the preceding segment of the trajectory is aligned with the velocity direction.

\section{Experiments}

More experimental details can be found in the accompanying video provided with this paper.

\subsection{Benchmark Comparisons}

\begin{figure*}[h]
    \centering
    \includegraphics[width=1.0\textwidth]{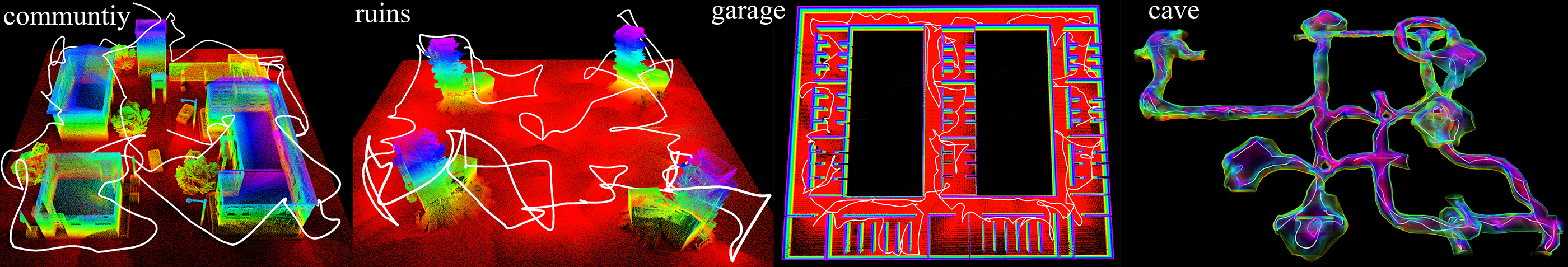}
    \caption{The trajectories and point cloud of the proposed method}
    \label{Fig6}
\end{figure*}

\begin{figure*}[h]
    \centering
    \includegraphics[width=1.0\textwidth]{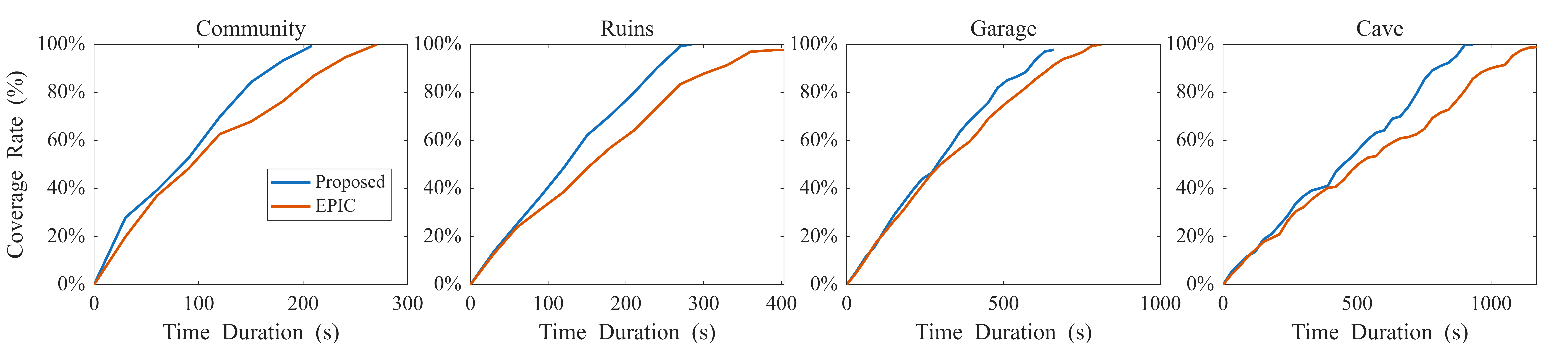}
    \caption{The exploration progress of state-of-the-arts benchmarks and the proposed method}
    \label{Fig7}
\end{figure*}

\begin{table}[h]
\centering
\caption{Results of Benchmark Comparisons}
\label{benchmark comparison}
\renewcommand{\arraystretch}{1.15}

\begin{tabular}{c c c c c}
\hline\hline
Scene & Method & Time. (s) & Length. (m) & Avg. Vel. (m/s) \\
\hline

Community
 & \textbf{Proposed} & \textbf{216.70}      & \textbf{619.33} & 2.86  \\
 & Epic   & 268.29  & 746.25 & 2.78 \\
\hline

Ruins
 & \textbf{Proposed} & \textbf{278.73}     & \textbf{1123.09} & \textbf{4.03}  \\
 & Epic   & 437.78  & 1323.57 & 3.01 \\
\hline

Garage
 & \textbf{Proposed} & \textbf{657.85}     & 2193.53 & 3.33  \\
 & Epic\cite{c10}   & 716.90  & 2222.39 & 3.10 \\
\hline

Cave
 & \textbf{Proposed} & \textbf{934.58}     & 3429.21 & \textbf{3.67}  \\
 & Epic\cite{c10}   & 1288.20 & 3452.38 & 2.68 \\
\hline\hline
\end{tabular}
\end{table}

In this work, four scenes are chosen to evaluate the proposed method against the state-of-the-art method, including two self-built scenes (ruins, scenes). To ensure fairness in comparison, two open-source scenes from EPIC\cite{c10} (garage, cave) are also chosen and used as baselines for evaluation. To ensure the quality of exploration, the point cloud data (PCD) files of scenes are divided into standard voxels and compared with the point clouds generated after the exploration. Under the condition of achieving comparable coverage rates, the performance of the two algorithms was evaluated.

The community represents an outdoor open environment of $[66 \times 66 \times 20]m^3$, composed of numerous buildings and trees. This scene is characterized by a relatively continuous and regular surface geometry, although there are some convex corners and abrupt changes. The ruin scene is another outdoor scene of $[100 \times 100 \times 30]m^3$. It consists of dilapidated structures with highly irregular and discontinuous surfaces. 

Each method was tested four times per scene. The maximum velocity was uniformly set to 5m/s for both methods. All tests were performed on an Intel Core i9-14900KF CPU. The simulation environment was kept consistent with EPIC by using MARSIM\cite{c27}. Table \ref{benchmark comparison} presents the results of exploration time, path length, and average velocity for the four best-performing runs, with entries exceeding a 10\% lead highlighted in bold. Fig. \ref{Fig6} shows the trajectory and point cloud of the proposed method. Fig. \ref{Fig7} compares the exploration progress of the proposed method and state-of-the-arts method.

In the community and ruins scene, the proposed method exhibited improved path topology and avoided redundant exploration of the same areas, thereby resulting in shorter trajectories and reduced exploration time. In addition, incorporating additional unknown frontiers can better guide the UAV to explore previously unobserved areas, particularly around complex corner regions. In the EPIC open-source garage and cave scenes, although the path lengths generated by the proposed algorithm were comparable to those of the baseline methods, the proposed approach achieved faster exploration times, with the improvement being particularly pronounced in the cave scene.

In terms of exploration progress, the proposed method consistently maintained a high exploration rate, whereas EPIC exhibited a gradual slowdown over time, as shown in Fig.\ref{Fig7}. This can be attributed to the hybrid frontier strategy, which provides better guidance toward unknown regions, thereby improving overall efficiency.

\subsection{Real-World Experiments}

\begin{figure}[h]
    \centering
    \includegraphics[width=0.45\textwidth]{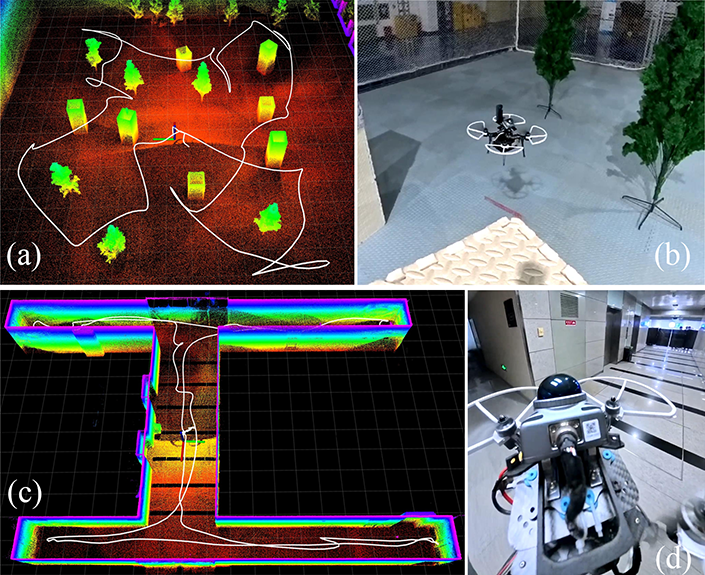}
    \caption{Performing Exploration Flight (a) Point clouds and trajectory of Column (b) third-person view (c) Point clouds and trajectory of Corridor (d) first-person view }
    \label{Fig10}
\end{figure}

\begin{table}[h]
\centering
\caption{Results of Real-World Experiments}
\label{real-world flight}
\renewcommand{\arraystretch}{1.15}

\begin{tabular}{c c c c c}
\hline\hline
Scene & Time. (s) & Length. (m) & Avg. Vel. (m/s) \\
\hline

Outdoor open area
 & 216.08  & 351.6 & 1.62 \\
\hline

Column
 & 117.06  & 92.83 & 0.79 \\
\hline

Corridor

 & 123.03  & 96.65 & 0.78 \\
\hline

\hline\hline
\end{tabular}
\end{table}

This work utilized a LiDAR-based quadrotor platform equipped with a Jetson Orin NX onboard computer, a Px4 autopilot and a Livox MID360 LiDAR mounted at a 30° pitch angle, as shown in Fig.\ref{Fig0}a. FAST-LIO2\cite{c26} is utilized to provide accurate odometry information. Additionally, an Insta360 panoramic camera was mounted on the UAV to capture a first-person perspective. However, it partially occludes the FOV of LiDAR. This also validates the necessity of self-calibration.

In this work, three scenes are selected for flight test:
\begin{enumerate}
\item Outdoor open area: $[64\times 80\times 16]m^3$.
\item Column: $[15\times 15\times 2.2]m^3$.
\item Corridor: $[13\times 33\times 2.2]m^3$.
\end{enumerate}

These scenes represent different topologies, sizes, and spatial environments, thereby demonstrating the robustness of the proposed algorithm. The outdoor open area is a scene with minimal obstacle-induced returns, used to validate the effectiveness of spherical projection in identifying free space regions. The corridor is a narrow (minimum width 0.8 m) and irregularly shaped environment, designed to evaluate both obstacle avoidance capability and the planning performance of the Hgrid under irregular terrain conditions. The column is a region characterized by an irregular, highly faceted surface with multiple corners.

Table. \ref{real-world flight} shows the results of real-world exploration. Fig.\ref{Fig0} and Fig.\ref{Fig10} represents the corresponding point clouds and trajectories. Fig.\ref{Fig0}b shows the first-person view of exploring. Fig.\ref{Fig0}d shows the frontiers, Hgrid and the planned path. Fig.\ref{Fig0}e represents the virtuall sphere for outwrad raycast. The flight speeds in simulation and real‑world experiments differ substantially. This discrepancy is mainly due to the heavier quadrotor, which exhibits reduced maneuverability and shorter endurance

In the outdoor open area, the UAV sequentially explored four buildings and conducted additional inspection of structures such as pavilions. Even though the majority of the airspace lacked horizontal LiDAR returns, the UAV still correctly labeled the free space by outward ray casting (Fig.\ref{Fig0}e) and flew safely.

In the column scene, the UAV captured high-quality point clouds of all surfaces of the columns and trees, without being adversely affected by surface irregularities or lack of smoothness. In the corridor scene, the UAV selected the most efficient exploration path and sequentially traversed the four corridors. The UAV also flew safely within narrow spaces of less than 0.8 m and completed the exploration.

\section{Conclusion}

In this work, SHIELD, Spherical-Projection Hybrid-Frontier Integration for Efficient LiDAR-based Drone Exploration, is presented. To avoid wrong ray-casting under big FOV, an observation-quality occupancy map is constructed. A hybrid frontier strategy is proposed to balance computational burden, exploration efficiency, and observation quality. An outward spherical-projection raycast method is proposed to label free without LiDAR returns in the occupancy map. Through the aforementioned innovations, this work successfully completed exploration in open areas lacking sufficient LiDAR returns and demonstrated advantages compared to the state-of-the-art method. Meanwhile, flight tests demonstrated that SHIELD maintained robust performance in narrow indoor environments, further validating its resilience under constrained real‑world conditions.

In the future, the method proposed in this paper will be extended to support multi‑UAV collaborative exploration, to further improve exploration efficiency.

\end{document}